\pdfoutput=1

\documentclass[11pt]{article}

\usepackage{acl}

\usepackage{times}
\usepackage{latexsym}
\usepackage{amsmath}
\usepackage{graphicx}
\usepackage{multirow}
\usepackage{booktabs}
\usepackage{adjustbox}
\usepackage{hyperref}

\usepackage[T1]{fontenc}

\usepackage[utf8]{inputenc}

\usepackage{microtype}

\title{EAVE: Efficient Product Attribute Value Extraction via \\ Lightweight Sparse-layer Interaction}

\author{Li Yang\thanks{Equal Contribution.}, \ Qifan Wang\textsuperscript{*}, Jianfeng Chi, Jiahao Liu, Jingang Wang\\ 
\textbf{Fuli Feng, Zenglin Xu, Lifu Huang, Dongfang Liu}\\
  \texttt{lyliyang@google.com, wqfcr@fb.com}}

\begin{document}
\maketitle
\begin{abstract}
Product attribute value extraction involves identifying the specific values associated with various attributes from a product profile. While existing methods often prioritize the development of effective models to improve extraction performance, there has been limited emphasis on extraction efficiency. However, in real-world scenarios, products are typically associated with multiple attributes, necessitating multiple extractions to obtain all corresponding values. 
In this work, we propose an Efficient product Attribute Value Extraction (EAVE) approach via lightweight sparse-layer interaction. Specifically, we employ a heavy encoder to separately encode the product context and attribute. The resulting non-interacting heavy representations of the context can be cached and reused for all attributes. Additionally, we introduce a light encoder to jointly encode the context and the attribute, facilitating lightweight interactions between them. To enrich the interaction within the lightweight encoder, we design a sparse-layer interaction module to fuse the non-interacting heavy representation into the lightweight encoder.
Comprehensive evaluation on two benchmarks demonstrate that our method achieves significant efficiency gains with neutral or marginal loss in performance when the context is long and number of attributes is large. Our code is available \href{https://anonymous.4open.science/r/EAVE-EA18}{here}.
\end{abstract}

\section{Introduction}
\begin{figure}
\begin{center}
\includegraphics[width=0.95\linewidth]{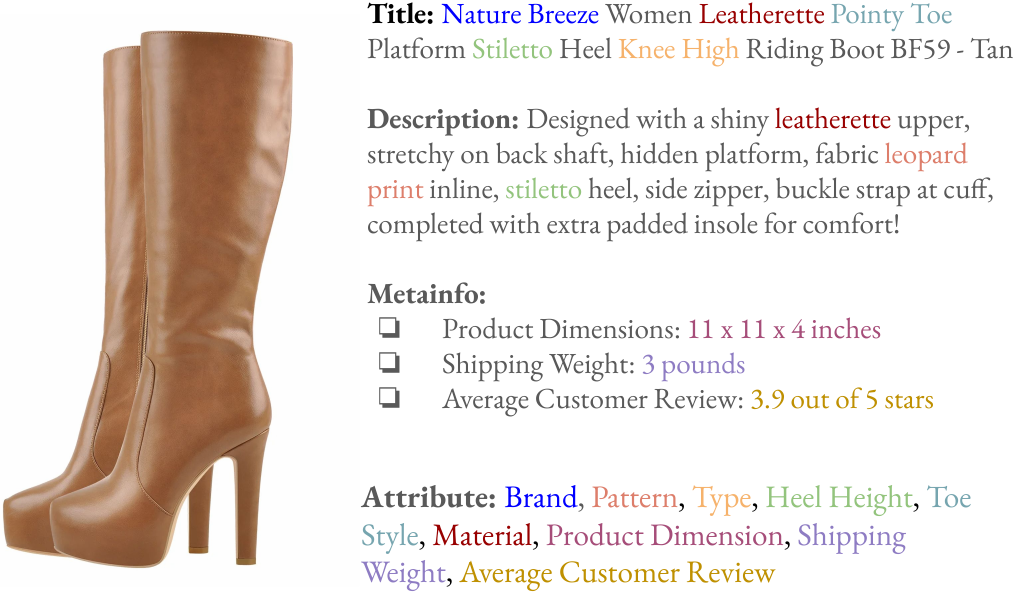}
\end{center}
\vspace{-4mm}
\caption{An example of product associated with multiple attributes and their corresponding values extracted from the product context.} \label{fig:product_example}
\vspace{-6mm}
\end{figure}

Product attributes serve as crucial features, carrying valuable information about a product. They constitute a fundamental aspect of e-commerce platforms, offering guidance to customers for product comparisons and purchase decisions. Additionally, product attributes play a vital role in various applications for merchants, including product recommendations~\cite{DBLP:conf/www/TruongZYLCPL22}, search~\cite{DBLP:conf/acl/NguyenRS20,DBLP:conf/naacl/LuHZWSY21}, and question answering systems~\cite{DBLP:conf/emnlp/ZhangDML20,DBLP:conf/naacl/RozenCMMZ21,huang-etal-2022-autoregressive}.
Attribute value extraction has attracted a lot of attention from both academia and industry, with a plethora of research~\cite{OpenTag,ScaleUpOpenTag,DBLP:conf/kdd/DongHKLLMXZZSDM20,AdaTag,DBLP:conf/acl/Shinzato0XC22,AMELI,ChenS0X23} being proposed to tackle this problem. 


With the advancements of Transformer models~\cite{Transformer,BERT}, attribute value extraction approaches based on Transformers~\cite{liyang2022mave,smartave} have achieved state-of-the-art performance. These methods concatenate the product attribute and context into a single text sequence and jointly encode it through the self-attention mechanism, effectively capturing the comprehensive interaction between the attribute and context. Despite their superior performance, they entail dense computation for extracting each attribute value individually. However, in real-world applications, efficient attribute value extraction is crucial for two main reasons. First, there are millions of new products being generated by the merchants everyday. Each product is typically associated with multiple attributes that describe its characteristics from various perspectives. For instance, as illustrated in Figure \ref{fig:product_example}, the `Shoes' has numerous attributes, such as `Brand', `Pattern', `Type', etc., necessitating multiple inferences and resulting in substantial computation costs. Second, attribute values are dynamic, undergoing changes such as updates to the product `Price' by the merchant. As a result, re-extraction is required whenever the product profile is updated. Hence, efficient extraction poses a significant research problem.

Several recent efficient sentence pair models~\cite{SentenceT5,VIRT,MixEncoder} can be applied to attribute value extraction. These models initially employ a dual encoder architecture to separately encode the product attribute and context. Subsequently, they utilize a late interaction layer to combine the attribute and context representations. This approach allows the product context to be cached and utilized for all attributes associated with the product, a concept explored in the context of attribute value extraction as well~\cite{ScaleUpOpenTag}. However, these late interaction techniques often yield less effective extraction results due to their neglect of the interaction during the attribute and context encoding. 
The interaction between the two encoders is particularly crucial in attribute value extraction, especially when the context length is short (additional discussion is presented in the experiments). In such cases, dense interactions become essential to thoroughly capture the connection between the attribute and context.

To address these challenges, in this paper, we propose a novel Efficient product Attribute Value Extraction (EAVE) approach via lightweight sparse-layer interaction. In particular, we employ a heavy encoder to separately encode the product context and attribute. The resulting non-interacting heavy representations of the context can be cached and reused for all attributes. Moreover, we introduce a light encoder to jointly encode the context and the attribute, enabling lightweight interactions between them. Furthermore, we design a sparse-layer interaction module to fuse the non-interacting heavy representation into the lightweight encoder, which further enriches the interactions between the context and the context. The evaluations on two product benchmarks demonstrate that our approach achieves similar performance to the state-of-the-art models while being much efficient. We summarize the main contributions as follows:
\begin{itemize}
\item We propose an efficient attribute value extraction method by introducing a heavy and a light encoder to learn effective product attribute-context representations. The heavy representations can be pre-computed and reused.
\item  We develop a sparse-layer interaction mechanism to fuse the non-interacting and interacting representations from the heavy and light encoders respectively to improve the model effectiveness with small computation cost. 
\item  We conduct extensive experiments and demonstrate the
effectiveness of the proposed approach over several state-of-the-art baselines.
\end{itemize}

\section{Related Work}

\textbf{Attribute Value Extraction}
Early attribute value extraction methods~\cite{10.1145/3209978.3210203, 10.1145/3292500.3330985}, including rule-based extraction~\cite{VandicDF12,GopalakrishnanIMRS12} and named entity recognition (NER)-based approaches~\cite{DBLP:conf/acl/BrookeHB16,10.1145/3331184.3331391}, suffer from limited coverage and closed-world assumptions. Various neural network models~\cite{BiLSTM-CRF,OpenTag,AdaTag} have also been introduced, formulating the extraction task as a sequential tagging problem. 
Recently, AVEQA~\cite{AVEQA} and MAVEQA~\cite{liyang2022mave} leverage the BERT architecture~\cite{BERT} by reformulating the problem as a question-answering task, establishing the state-of-the-art in attribute value extraction. Meanwhile, OA-Mine~\cite{OAMine} and MixPAVE~\cite{MixPave} focus on zero-shot and few-shot attribute value extraction. Notably, several multi-modal works~\cite{DBLP:conf/emnlp/TanB19,MJAVE,PAM} explore product visual features to enhance attribute value extraction. SMARTAVE~\cite{smartave} designs a structured multi-modal Transformer to better encode the correlation among different product modalities. Despite achieving state-of-the-art performance, these Transformer-based methods require significant computational resources for extracting values for billions of attributes, which can be resource-intensive in many real-world scenarios.

More recently, generation-based approaches~\cite{Generationbased} have been proposed, including those leveraging the LLMs~\cite{LLM1,LLM2,LLM3}, to directly decode the attribute and value pairs together, eliminating the requirement of the product taxonomy. However, generation-based models usually fail to generate a complete set of attribute-value pairs. Moreover, their performances are suboptimal compared to the extraction-based models. More discussions are presented in the experiments and appendix~\ref{app:llm}.

\begin{figure*}[t!]
    \centering
    \includegraphics[width=0.95\textwidth]{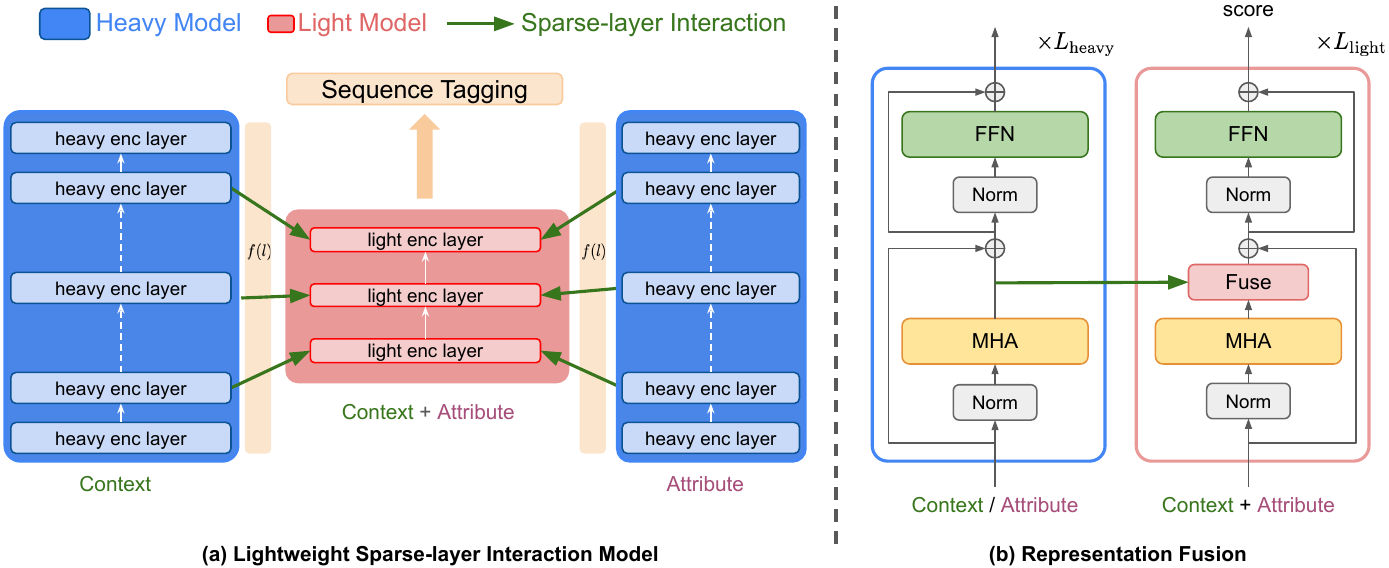}
    \vspace{-2mm}
    \caption{Overview of our EAVE model. (a) There are three key components: 1) Heavy model is used to encode the context and attribute, learning their non-interacting representations. 2) Light model generates the interacting representations of the concatenated context and attribute. 3) Sparse-layer interaction fuses the two representations from heavy and light encoders through sparse layer mapping (detailed in (b)).}
    \label{fig:model_architecture}
    \vspace{-5mm}
\end{figure*}

\noindent
\textbf{Efficient Text Pair Encoders}
Efficient text pair encoders~\cite{Sentence-BERT,DiPair,DeFormer,PolyEncoder,sun-etal-2023-fusion} can be employed in attribute value extraction by treating the product attribute and context as two text sequences. Most of these methods utilize a dual-encoder structure to individually encode the two text pieces, followed by a late interaction layer to combine the representations. In comparison with cross-attention models, they are more efficient as the text pair representations are computed independently without considering their interaction, allowing for caching and reuse. However, they often perform less optimally than cross-attention models. Several recent research~\cite{SentenceT5,VIRT,MixEncoder} has focused on enhancing the performance of dual-encoder models in text pair modeling tasks by incorporating text pair interactions while preserving the efficiency of dual encoders. 
While these methods generally improve efficiency, the interactions between dual encoders are not fully explored. Moreover, given the typically short lengths of attributes and context, interactions between text pairs become exceedingly important and thus require careful modeling.




\section{Method}

\subsection{Approach Overview}
The objective of attribute value extraction is to extract the corresponding value for each attribute from the product context, i.e., product title, description, etc. For instance, in Figure~\ref{fig:product_example}, the associated value for attribute `Brand' is `Nature Breeze'. 
Previous state-of-the-art methods often concatenate the context and the attribute into a single sequence, and feeds into a Transformer encoder. Given that the context is usually much longer than the attribute, the input sequences for different attributes only vary by a small portion. However, each of these similar inputs undergoes a forward pass through the same Transformer model, a process we consider inefficient and believe can be optimized.

The overall model architecture of EAVE is shown in Figure~\ref{fig:model_architecture}. Essentially, our model contains three major components: (1) A heavy Transformer encoder to learn representations for context and attribute independently, which is referred to non-interacting representations since there is no interaction between context and attribute. These representations can then be pre-computed and cashed after the training. (2) A light Transformer encoder to compute representations for context and attribute jointly, allowing full attention between them. We referred these to interacting representations. (3) A sparse-layer interaction module to fuse the non-interacting and interacting representations and enhance the final representation. A sequence tagging module is employed on the final representation for the attribute value extraction.

\subsection{Heavy Encoder}
The inefficiency in cross-attention extraction models~\cite{smartave,MixPave} stems from the interdependence of computations for context and attribute representations, coupled with their interactions within a single model. Given that the context sequence is typically much longer than the attribute, the entire representation undergoes certain perturbation due to the influence of attribute tokens.
However, the representation computation, including the interaction between context and attribute, is dominated by the context part. Consequently, we propose isolating these interactions into a lightweight model, allowing the other representations to be independently computed through a heavy encoder. This approach enables the pre-computation and caching of the context representation, which can be used for multiple attribute extractions. Formally, the computation of each heavy layer's self-attention is as follows:
\begin{equation}
\label{eq:heavy_mdoel}
\small
\begin{split}
    y'^p_l &= \text{SelfAttn}(\text{Norm}(x^p_{l-1})), \ \ y^p_l = y'^p_l + x^p_{l-1} \\
    x'^p_l &= \text{MLP}(\text{Norm}(y^p_l)), \ \ x^p_l = x'^p_l + y^p_l
\end{split}
\end{equation}
where $p \in \{c, a\}$ is the superscript for context or attribute. $x^p_l$ is the input heavy encoder representation for layer $l$. $y'^p_l$ is the self-attention output representation in layer $l$, which will be extracted and fused into the light Transformer encoder later. In this way, both the non-interaction representations for context ($y'^c_l$) and attribute ($y'^a_l$) can be pre-computed and cached. We initialize the heavy encoder from a pretrained T5~\cite{2020t5}.

\subsection{Light Encoder}
The heavy encoder learns the context representation independently, allowing its computation to be cashed and reused for various attribute value extraction. Previous works~\cite{AVEQA,OAMine} have demonstrated that the interaction between the context and attribute is critical for improving the effectiveness of the extraction.
Therefore, in this work, we first introduce a light Transformer encoder to facilitate the lightweight interaction, and then fuse the representations from the heavy encoder into the light interacting representations through sparse-layer interaction (in Section~\ref{sec:fuse}). Specifically, the light encoder has a similar architecture as the heavy encoder, but with a much smaller number of parameters in terms of hidden size, number of attention heads and number of layers. The input to the light encoder is the concatenation of the context and attribute tokens. We initialize the light encoder from a smaller pretrained T5~\cite{2020t5} checkpoint comparing with heavy encoder.

\subsection{Sparse-layer Interaction}
\label{sec:fuse}
In the attribute value extraction task, the interactions between the product context and attribute are crucial, especially when the context length is short. While the interactions captured in the light Transformer encoder are efficient, they may not be as comprehensive for effective extraction when compared to methods using a heavy encoder with full attention. To enhance effectiveness, we introduce a sparse-layer interaction approach to merge the non-interacting representations from the heavy model into the interacting light encoder representations. This enables the transfer of valuable information encoded in the dense representations, facilitating interaction with the lightweight representation. The overall extraction performance is improved with only a marginal increase in computation cost.

Concretely, since the heavy encoder has a large number of layers, we first identify a sparse set of layers in the heavy encoder, through a layer mapping function $f$, to be fused into the light encoder. In this paper, we choose an even distribution layer mapping, i.e. every other $L_\text{heavy} / L_\text{light}$ heavy encoder layer will be mapped to a light encoder layer (more ablation studies on different layer mapping schemes are presented in Section~\ref{sec:layer_mapping}). For each layer $l$ in the light encoder, we then fuse representations from the heavy encoder $y'^c_{f(l)}$ and $y'^a_{f(l)}$ into the light encoder at the same location where the heavy encoder representations are extracted:
\begin{equation}
\label{eq:light_model}
\small
\begin{split}
    y'_l &= \text{SelfAttn}(\text{Norm}(x_{l-1})) \\
    y_l &= \text{Fuse}(y'_l, y'^c_{f(l)}, y'^a_{f(l)}) + x_{l-1} \\
    x'_l &= \text{MLP}(\text{Norm}(y'_l)) \\
    x_l &= x'_l + y_l
\end{split}
\end{equation}
where $x_l$ is the input light encoder representation for layer $l$. $y'_l$ is the self-attention output representation in layer $l$, which will be fused with the heavy encoder representations $y'^c_{f(l)}$ and $y'^a_{f(l)}$ before skip connection. Note that we always fuse light and heavy representations extracted from the same location in Transformer encoder to ensure efficient fusion. As shown in Figure~\ref{fig:model_architecture}(b), we select the location to be immediately after the self-attention layer and before its skip connection (different locations for representation extraction and fusion are provided in Section~\ref{sec:fuse_location}). The heavy encoder representations and light encoder representations are fused with a linear interpolation for all layers.
\begin{equation}
\small
\begin{split}
&\text{Fuse}(y'_l, y'^c_{f(l)}, y'^q_{f(l)}) \\ 
= \;&(1 - \alpha) * y'_l + \alpha * W_\text{adp}\text{Concat}(y'^c_{f(l)}, 
 y'^q_{f(l)})
\end{split}
\end{equation}
where we first concatenate  $y'^c_{f(l)}$ and $y'^a_{f(l)}$ along the sequence dimension, followed by an linear adaptor, $W_\text{adp}$, to project the hidden size into the same size as the light encoder, then linearly fuse with light representations $y'_l$. We also experiment with other fusion functions and provide discussion in Section~\ref{sec:fusion_method}. The final fused output of the light Transformer encoder is fed into the sequence tagging module for extraction.

\subsection{Discussion}
In real-world attribute value extraction, we only need to invoke the heavy encoder on each product context and attribute once to pre-compute the non-interacting representations. For each attribute, the light encoder and the sparse-layer interaction are used to compute interacting representations for the context-attribute pair. The total computation cost for each product with $N$ attributes can be computed as $C^c_{h} + N\cdot (C^a_h + C_{l} + C_{sli})$, while previous methods would require $N\cdot C^{c+a}_h$,
where $C^c_{h}$, $C^a_{h}$ and $C^{c+a}_{h}$ are the computation cost of the heavy encoder on the context, attribute and their concatenation, respectively. Typically $C^a_h$ is much smaller than $C^c_{h}$, since attribute is much shorter than context. $C_{l}$ is the cost of the light encoder on concatenated context and attribute. $C_{sli}$ is the cost of the fusion module. It can be seen that our approach is much efficient. It is worth mentioning that if ignoring the $C^c_{h}$ and $C^a_h$ (since they can be cashed), our method becomes even more efficient compared to those cross-attention heavy encoder methods~\cite{smartave,MixPave}.
\begin{table*}[t!]
\begin{adjustbox}{width=0.57\width,center}
\begin{tabular}{c|ccc|ccc|ccc|ccc|ccc|c}
\toprule
\textbf{MAVE}       & \multicolumn{3}{|c}{\textbf{Season}}                             & \multicolumn{3}{|c}{\textbf{Department}}                & \multicolumn{3}{|c}{\textbf{Resolution}}                & \multicolumn{3}{|c}{\textbf{Compatibility}}             & \multicolumn{3}{|c|}{\textbf{All}}                       & \multirow{2}{*}{\textbf{GFLOPS}}                     \\
\cmidrule{1-16}
\textbf{Models} & \textbf{${\bf P}$(\%)} & \textbf{${\bf R}$(\%)} & \textbf{${\bf F_1}$(\%)} & \textbf{${\bf P}$(\%)} & \textbf{${\bf R}$(\%)} & \textbf{${\bf F_1}$(\%)} & \textbf{${\bf P}$(\%)} & \textbf{${\bf R}$(\%)} & \textbf{${\bf F_1}$(\%)} & \textbf{${\bf P}$(\%)} & \textbf{${\bf R}$(\%)} & \textbf{${\bf F_1}$(\%)} & \textbf{${\bf P}$(\%)} & \textbf{${\bf R}$(\%)} & \textbf{${\bf F_1}$(\%)} &  \\
\midrule
DiPair~\cite{DiPair}       & 88.35           & 88.98           & 88.53              & 90.78           & 92.09           & 91.43             & 93.16           & 94.34           & 93.72              & 95.78           & 96.56           & 96.21              & 92.84           & 94.47           & 93.61              & 35.18              \\
VIRT~\cite{VIRT}       & 89.76           & 90.81           & 90.34              & 93.48           & 94.82           & 93.78             & 94.31           & 95.72           & 95.47              & 97.18         & 98.12          & 97.56              & 94.77           & 96.39           & 95.52             & 41.67              \\
MixEncoder~\cite{MixEncoder}       & 89.88           & 90.92           & 90.47              & 93.18          & 94.90           & 93.52             & 95.14           & 96.17           & 95.62              & 97.75         & 98.29          & 98.03              & 94.92           & 96.76           & 95.64              & 33.93              \\
\midrule
Generation~\cite{Generationbased}       & 89.51           & 87.18           & 88.33              & 91.38           & 89.47           & 90.41              & 94.86           & 91.69           & 93.25              & 96.25           & 95.85           & 96.05              & 94.31 & 92.48 & 93.52              & 87.57              \\
AVEQA~\cite{AVEQA}       & 91.63           & 93.82           & 92.71              & 96.51           & 97.03           & 96.77              & 98.20           & 98.79           & 98.49              & 99.83           & 99.92           & 99.88              & 97.96           & 98.44           & 98.20              & 402.47              \\
MAVEQA~\cite{liyang2022mave}       & 91.15           & 94.13           & 92.62              & 96.39           & 97.35           & 96.87              & 98.07           & 98.81           & 98.44              & 99.56          & 99.81          & 99.68              & 97.85           & 98.57           & 98.21              & 446.24              \\
SMARTAVE~\cite{smartave}       & 91.17           & 94.16          & 92.64              & 96.47           & 97.38           & 96.92              & 98.15           & 98.78           & 98.46              & 99.62          & 99.83          & 99.72              & 97.88           & 98.60           & 98.24              & 465.71              \\
\midrule
EAVE       & 92.35           & 94.94           & 93.63              & 97.05           & 97.84           & 97.44              & 97.78           & 98.25           & 98.01              & 99.83           & 99.67           & 99.75              & 97.94           & 98.03           & 97.98              & 42.46    \\
\bottomrule
\end{tabular}
\end{adjustbox}
\vspace{-2mm}

\quad

\begin{adjustbox}{width=0.6\width,center}
\begin{tabular}{c|ccc|ccc|ccc|ccc|c}
\toprule
\textbf{AE-110K} & \multicolumn{3}{|c}{\textbf{Brand Name}} & \multicolumn{3}{|c}{\textbf{Material}} & \multicolumn{3}{|c}{\textbf{Pattern Type}} & \multicolumn{3}{|c|}{\textbf{All}}     & \multirow{2}{*}{\textbf{GFLOPS}}  \\
\cmidrule{1-13}
\textbf{Model} & \textbf{${\bf P}$(\%)} & \textbf{${\bf R}$(\%)} & \textbf{${\bf F_1}$(\%)} & \textbf{${\bf P}$(\%)} & \textbf{${\bf R}$(\%)} & \textbf{${\bf F_1}$(\%)} & \textbf{${\bf P}$(\%)} & \textbf{${\bf R}$(\%)} & \textbf{${\bf F_1}$(\%)} & \textbf{${\bf P}$(\%)} & \textbf{${\bf R}$(\%)} & \textbf{${\bf F_1}$(\%)} & \\
\midrule
DiPair~\cite{DiPair}     & 87.64 & 89.87 & 89.25 & 76.31 & 80.29 & 78.62    
& 78.72    & 82.58    & 80.43      & 79.74  & 70.69  & 76.82    & 3.75           \\
VIRT~\cite{VIRT}              & 90.56 & 91.90 & 91.33 & 79.29 & 81.75 & 80.28    
& 81.07    & 84.83    & 82.81      & 77.86  & 80.27  & 78.95    & 4.92   \\
MixEncoder~\cite{MixEncoder}  & 90.68 & 92.55 & 91.41 & 80.40 & 81.97 & 81.24    
& 81.86    & 84.98    & 83.15      & 78.53  & 80.74  & 79.21    & 3.42             \\
\midrule
Generation~\cite{Generationbased}       & 90.18           & 88.65           & 89.41              & 79.83           & 78.75           & 79.29             & 80.26           & 81.47           & 80.86              & 79.62           & 75.54           & 77.53 & 6.68              \\
AVEQA~\cite{AVEQA}            & 96.36 & 98.57 & 97.46 & 84.01 & 88.89 & 86.38    & 87.42    & 90.41    & 88.89      & 85.01  & 86.09  & 85.54     & 16.11      \\
MAVEQA~\cite{liyang2022mave}  & 96.21 & 98.52 & 97.39 & 83.96 & 88.65 & 86.21    & 87.55    & 90.57    & 89.03      & 84.96  & 86.05  & 85.51     & 18.54      \\
SMARTAVE~\cite{smartave}      & 96.32 & 98.59 & 97.43 & 84.01 & 88.65 & 86.27    & 87.48    & 90.49    & 88.96      & 85.12  & 86.07  & 85.49     & 19.75               \\
\midrule
EAVE          & 96.90      & 97.13     & 97.02       & 53.27    & 53.95    & 53.61      & 87.21    & 91.10     & 89.11      & 85.01  & 84.24  & 84.62     & 5.20      \\
\bottomrule
\end{tabular}
\end{adjustbox}
\vspace{-2mm}
\caption{Performance comparison of selected attributes on both MAVE and AE-110K datasets. For all efficient methods, the GFLOPS do not include the pre-computation time. The standard deviation of our model on the $F_1$ metric is 0.13, indicating the statistical significant of the results.}
\label{tab:mave_ae110_metrics}
\vspace{-5mm}
\end{table*}

\section{Experiments}
\subsection{Datasets}

\noindent\textbf{MAVE\footnote{\url{https://github.com/google-research-datasets/MAVE}}}~\cite{liyang2022mave} is a large and diverse dataset for product attribute extraction study, which contains 3 million attribute value annotations across 1257 fine-grained categories created from 2.2 million cleaned Amazon product profiles \cite{AmazonReview}. We use product title, description and metadata as context. We randomly split product ids into train and eval sets by 9:1.

\noindent\textbf{AE-110K\footnote{\url{https://raw.githubusercontent.com/lanmanok/ACL19_Scaling_Up_Open_Tagging/master/publish_data.txt}}}~\cite{ScaleUpOpenTag} is collected from AliExpress Sports \& Entertainment category, which contains over 110K data examples with more than 2.7K unique attributes and 10K unique values. The context length in MAVE is much longer then that in AE-110K. More details are in the Appendix.

\subsection{Baselines and Settings}
Our model is compared with seven state-of-the-art baselines, including three Transformer-based extraction models, AVEQA~\cite{AVEQA}, MAVEQA~\cite{liyang2022mave}, and SMARTAVE~\cite{smartave}, one generation-based model~\cite{Generationbased} and three efficient text pair methods, DiPair~\cite{DiPair}, VIRT~\cite{VIRT} and MixEncoder~\cite{MixEncoder}. 

For all models in this paper, we use a learning rate of 1e-5, batch size 128, and trained up to 200k steps on 16 Cloud TPU V5 devices through data parallelism. Adam \cite{Kingma2014AdamAM} optimizer is used during training. For MAVE, we truncate the context sequence length to 512, and set attribute sequence length to 32. A T5-large model is used as the heavy encoder. For AE-110K, we pad or truncate the context sequence length to 128, and pad or truncate the attribute sequence length to 16. A T5-base model is used as the heavy encoder. For SMARTAVE, we use the SMARTAVE-text since we only focus on text based extraction. For generation-based method, we use beam search with width 4 and report the `common first' result (which is the best). For the efficient text pair methods, a same sequence tagging module is applied to conduct the final extraction.
More details on the hyper-parameters are presented in Appendix \ref{training_details}.


\section{Main Results}
We present precision, recall, F1, and GFLOPS per example metrics for selected attributes, along with overall results on both MAVE and AE-110K in Table~\ref{tab:mave_ae110_metrics}. Several key observations can be derived from these results. \textbf{First}, in comparison to state-of-the-art Transformer-based attribute value extraction methods, our EAVE approach achieves a comparable overall performance, with only a 0.2\% and 0.9\% decrease in F1 on MAVE and AE-110K, respectively. Notably, it significantly improves efficiency by a factor of 10, showcasing the effectiveness of our approach in efficient attribute value extraction.
\textbf{Second}, although EAVE entails slightly higher computational costs compared to other efficient text pair methods, it substantially enhances extraction performance. For instance, the overall F1 score of EAVE increases by 2.34\% and 5.41\% on MAVE and AE-110K, respectively, compared to MixEncoder. We hypothesize that our specially designed lightweight model and sparse-layer interaction are more adept at capturing interactions compared to the late interaction layer used in previous methods. \textbf{Third}, another interesting observation is the lower overall performance on AE-110K compared to MAVE. This discrepancy is attributed to the product context in AE-110K being solely derived from the title, which is very short. The context alone may not provide sufficient useful contextual information for the attribute value extraction task. Instead, the context-attribute interaction is deemed more crucial than the context itself, aligning with our expectations. \textbf{Fourth}, the generation-based model does not perform well compared to the extraction-based models, especially on AE-110K. Our hypothese is that for the attribute value extraction task, in most cases, the value is from a text span in the product context and thus extractive models are more effective compared to the generative models. The observation is consistent with the findings in \cite{LLM1} that a smaller LLM like Beluga-7B utilizing in-context learning fails to outperform a fine-tuned extraction-based model (i.e., AVEQA) with a significantly smaller size. More LLM-based results are reported in Appendix \ref{app:llm}.
\begin{figure}[t]
    \centering
    \includegraphics[width=0.49\linewidth]{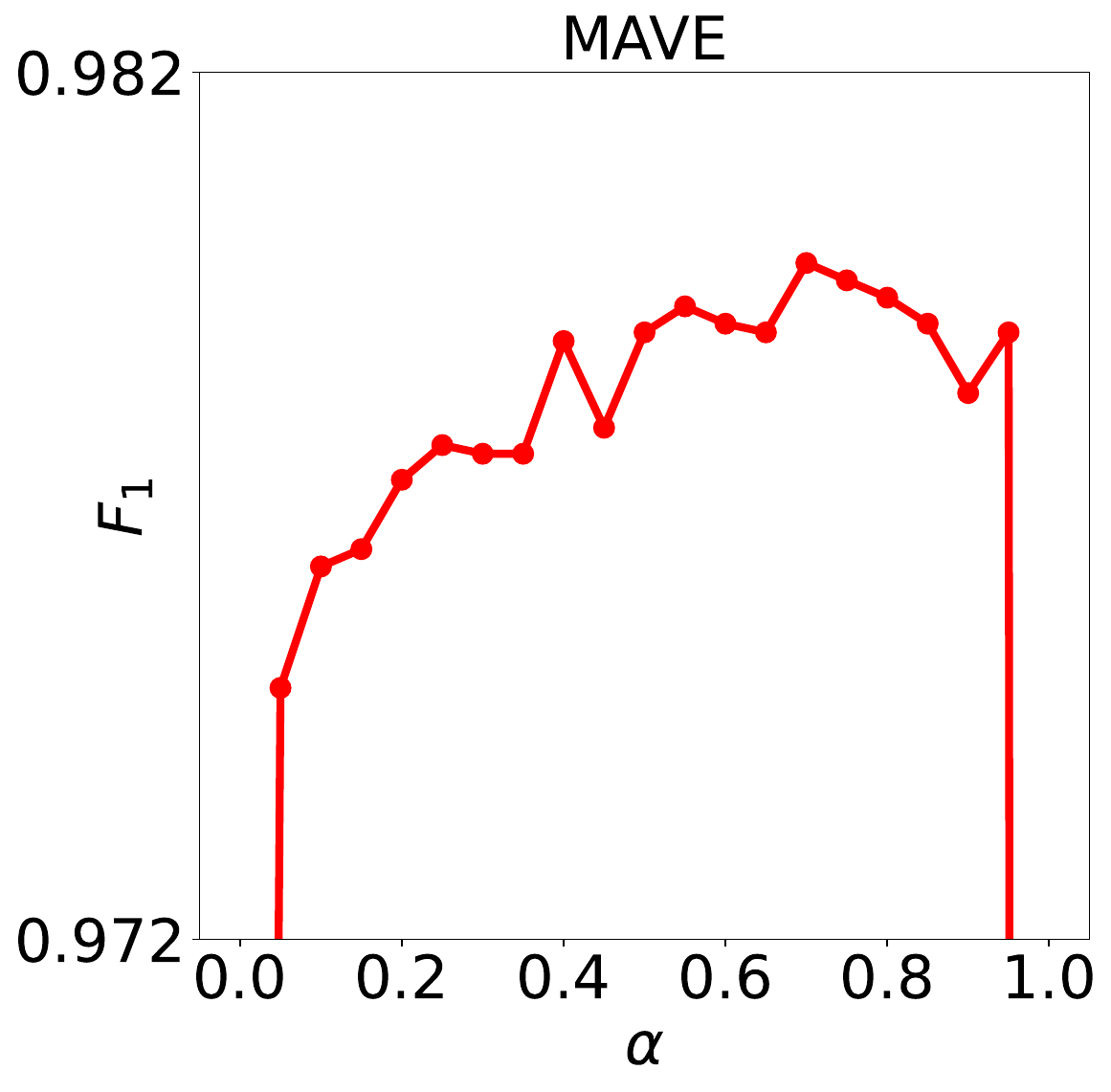}
    \includegraphics[width=0.49\linewidth]{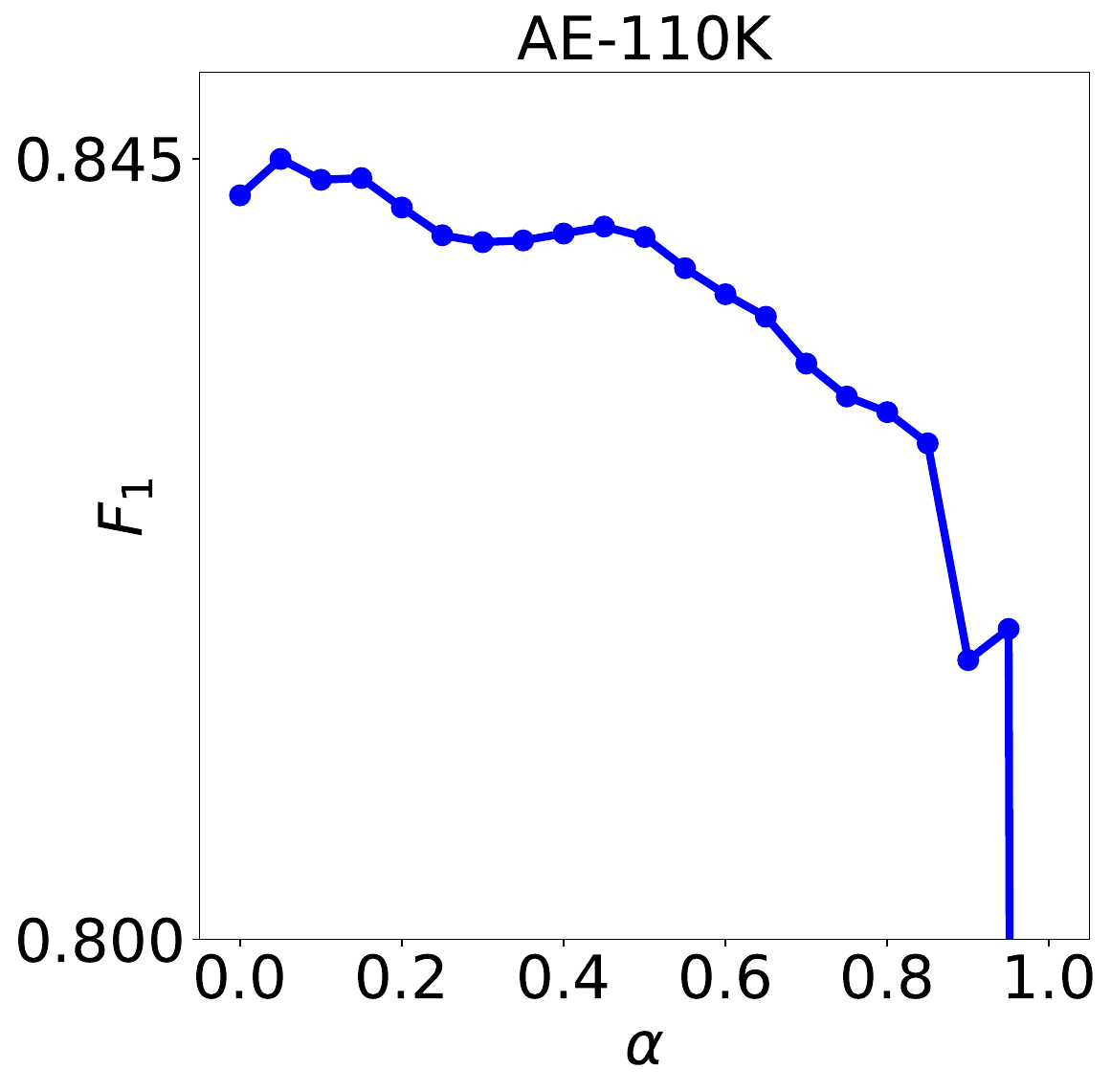}
    \vspace{-6mm}
    \caption{Impact of the sparse-layer interaction by varying the fusion weight $\alpha$ on both dataset.}
    \label{fig:metrics_alpha}
    \vspace{-5mm}
\end{figure}
\section{Analysis and Discussion}
\label{sec:analysis}

\subsection{Impact of the Sparse-layer Interaction}
\label{sec:fusion_method}
The sparse-layer interaction is a key component in our approach. To assess the impact of this module, we conducted an experiment by varying the value of $\alpha$ from 0 to 1. The evaluation results of EAVE for different values of $\alpha$ are illustrated in Figure~\ref{fig:metrics_alpha}.
It can be seen that when completely removing the heavy encoder representations by setting $\alpha$ to 0, and the model degrades to a single-light encoder baseline, which does not perform well compared to the heavy Transformer model. Conversely, when $\alpha$ is set to 1, the light encoder's self-attention outputs are entirely dominated by the heavy representations, resulting in the absence of interaction between context and attribute, which unsurprisingly yields suboptimal results. We also observe from the results that different datasets might have different optimal values of $\alpha$ for merging the non-interacting and interacting representations.

\subsection{Different Fusion Methods}
The fusion function in the sparse-layer interaction is an important factor that would impact the model performance.
There are various fusion methods that can be used. In this study, we experiment with three representation fusion algorithms. The simplest one is using a linear interpolation with a fixed weight $\alpha$ for all layers, which is the one described and used in our main experiments. The second algorithm is the learned $alpha$. Specifically, instead of using a fixed $\alpha$ for all layers, we use a learnable zero-initialized $\alpha$ for each layer. We expect the light encoder first learns context-attribute interaction without the influence from non-interacting heavy representations, then as $\alpha$ updated to a non-zero value, the light encoder will gradually adapt to non-interacting  heavy representations.
The third fusion choice is the cross-attention fusion by first concatenating and project non-interacting heavy representations, and then updating the interacting light representations via cross attending to the projected non-interacting  heavy representations, followed by a skip connection as follows: 
\begin{equation}
\small
\begin{split}
&\text{Fuse}(y'_l, y'^c_{f(l)}, y'^q_{f(l)}) \\ 
= \;&\text{CrossAttn}(y'_l, W_\text{adp}\text{Concat}(y'^c_{f(l)}, y'^q_{f(l)}) + y'_l
\end{split}
\end{equation}
In contrast to linear interpolation combination, in this setup, a token's light representation not only sees heavy representation of the token itself, but also sees heavy representations of other tokens. 

The results on MAVE and AE-110K for different fusion methods are shown in Table~\ref{tab:fusion_method}. It can be seen that our method achieves the best results by grid searching to select the best fixed $\alpha$. It is also clear that while the strategy of learnable $\alpha$ achieves similar or slight worse performance, the cross-attention method performs not as good as other methods.

\begin{table}[t]
\begin{adjustbox}{width=0.65\width,center}
\centering
\begin{tabular}{ccccccc}
\toprule
    & \multicolumn{3}{c}{\textbf{MAVE}} & \multicolumn{3}{c}{\textbf{AE-110K}} \\
\midrule
\textbf{Fusion method}           & \textbf{$P$ (\%)} & \textbf{$R$ (\%)} & \textbf{$F_1$ (\%)} & \textbf{$P$ (\%)} & \textbf{$R$ (\%)} & \textbf{$F_1$ (\%)} \\
\midrule
Fixed $\alpha$  & 97.94                 & 98.03              & 97.98         & 85.01                 & 84.24              & 84.62 \\
Learned  $\alpha$ & 97.02                 & 97.08              & 97.05     & 84.31                 & 84.24              & 84.27     \\
Cross-attn                   & 95.85                 & 96.76              & 96.30  & 80.81                 & 76.31              & 78.50 \\
\bottomrule
\end{tabular}
\end{adjustbox}
\vspace{-2mm}
\caption{Ablation study of representation fusion methods. For interpolation with fixed $\alpha$, we use $\alpha=0.7$ for MAVE and $\alpha=0.05$ for AE-110K.}
\label{tab:fusion_method}
\vspace{-5mm}
\end{table}


\subsection{Impact of Heavy and Light Encoders}
\begin{figure}[t!]
    \centering
    \includegraphics[width=0.49\linewidth]{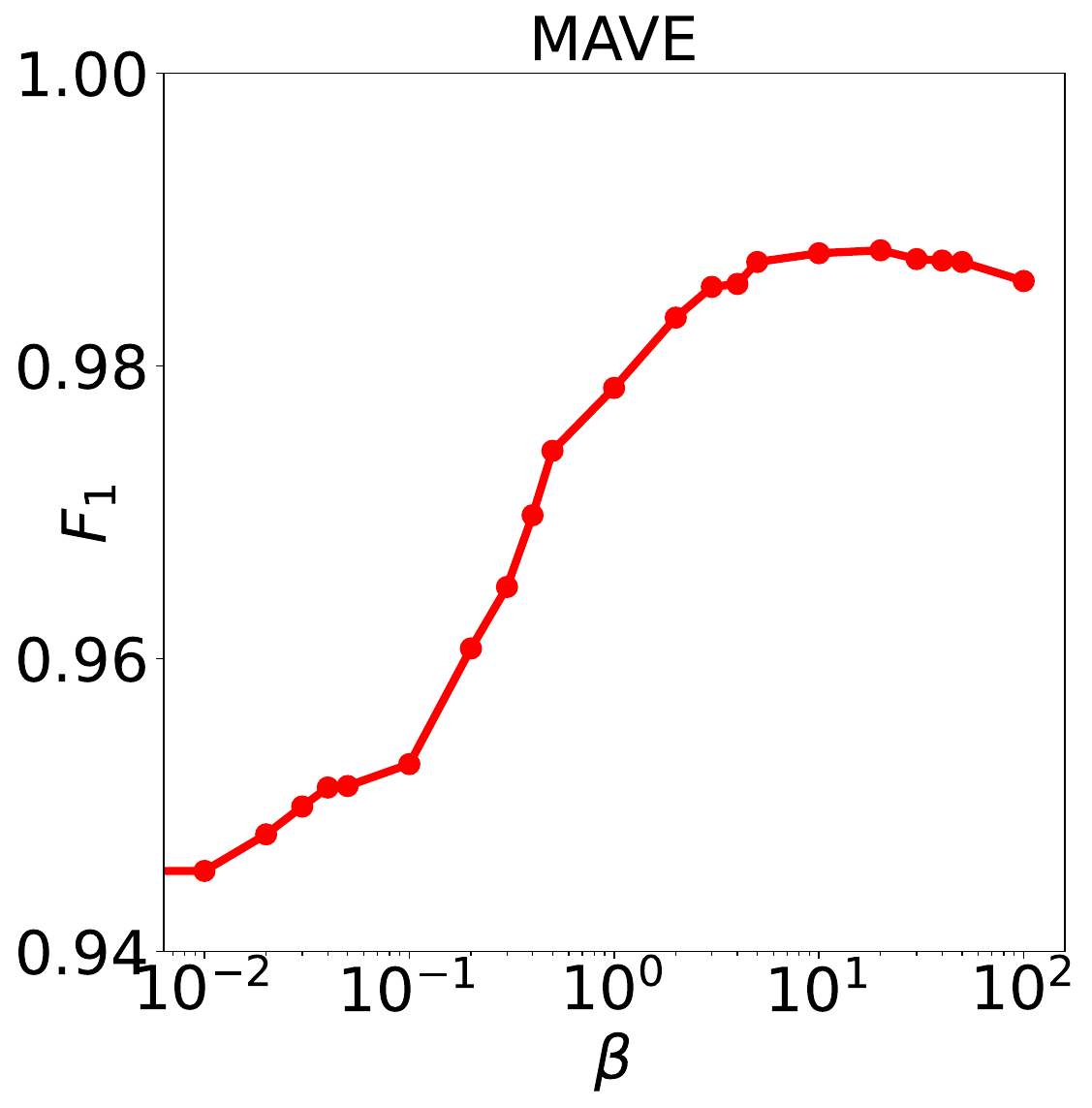}
    \includegraphics[width=0.49\linewidth]{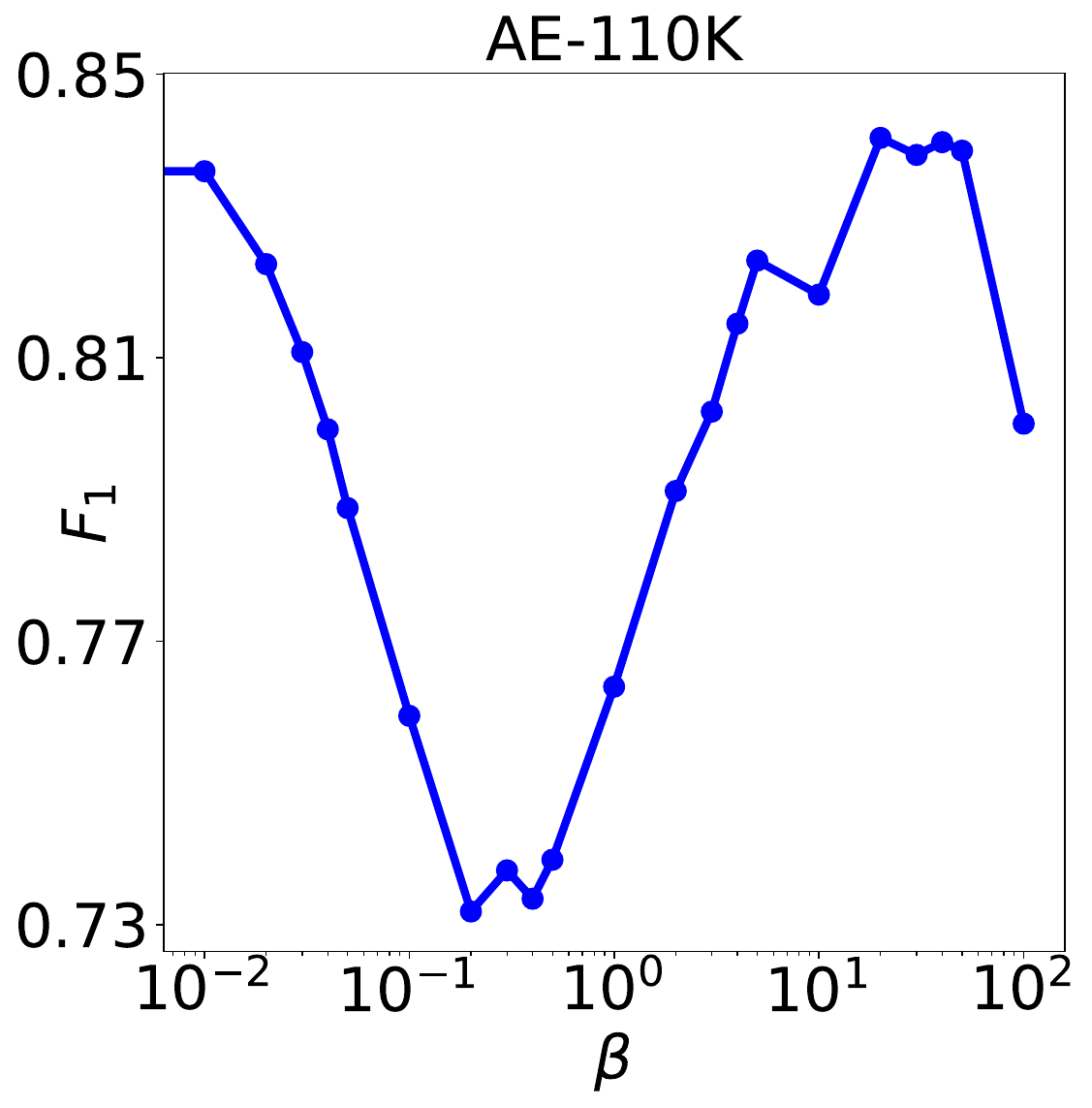}
    \vspace{-6mm}
    \caption{Impact of the heavy-light learning rate ratio $\beta$ on the MAVE and AE-110K dataset. For MAVE, $\alpha$ is fixed to 0.7. For AE-110K, $\alpha$ is fixed to 0.05.}
    \vspace{-5mm}
    \label{fig:metrics_beta}
\end{figure}

There are two separate encoders for learning the non-interacting representations and interacting representations. It is useful to understand how they would affect the model performance. To investigate this, we conduct experiments by modifying their parameter learning rates. In particular, we use a fixed learning rate $\text{LR}_\text{light}$ for the light encoder but varying the heavy encoder learning rate as $\text{LR}_\text{heavy} = \beta \text{LR}_\text{light}$, where $\beta \ge 0$ is the learning rate ratio. Figure~\ref{fig:metrics_beta} shows the performance on the MAVE and the AE-110K datasets as a function of $\beta$. It can be seen that, the performance on MAVE gets better as $\beta$ increases from 0, suggesting the importance of updating the heavy encoder for non-interacting representations. This further validates our hypothesis that long context alone contains useful information for the attribute value extraction task. On the other hand, for AE-110K, the performance decreases as the value of $\beta$ goes up, and increases again as $\beta$ pass around 1.0, indicating that heavy encoder itself does not provide much useful information when context is short. In this case, interaction between context and attribute are more important.

\begin{figure}[h!]
    \centering
    \includegraphics[width=0.95\linewidth]{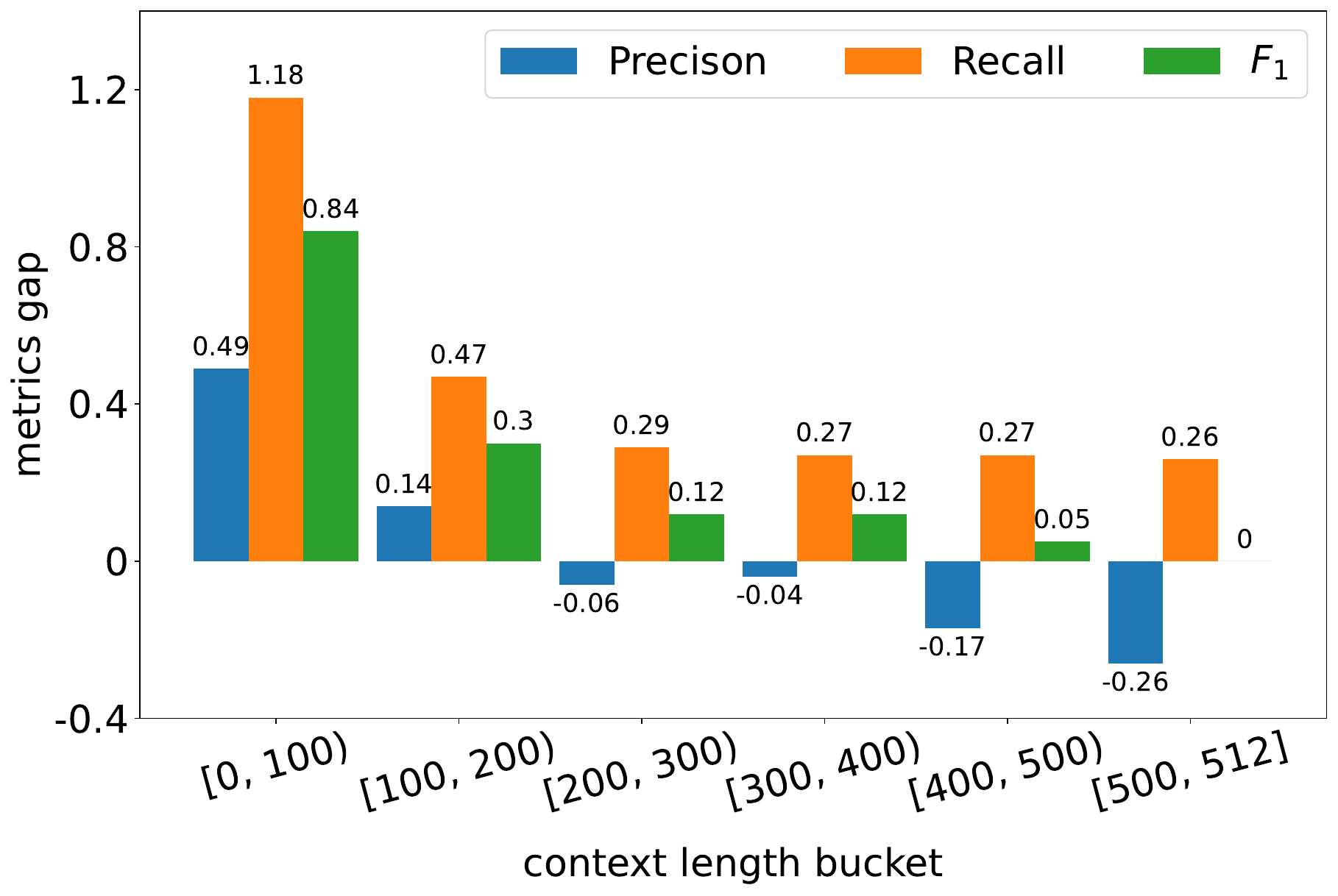}
    \vspace{-3mm}
    \caption{Precision, recall, and F1 \textbf{gaps} from the EAVE to the Transformer baseline on the MAVE dataset. We use a fixed $\alpha=0.7$ and $\beta=1.0$.}
    \vspace{-5mm}
    \label{fig:mave_context_length}
\end{figure}
\subsection{Results on Different Context Length}
\label{sec:context_length}
To understand how our approach behaves on different context lengths, we conduct an experiment on MAVE and report the evaluation metrics on different groups with various context sequence lengths from $\{0-100, 100-200, 200-300, 300-400, 400-500, 500-512\}$. The results are shown in Figure~\ref{fig:mave_context_length}. It can be seen from the results that with the increasing of the context length, the performance gaps between our model and the Transformer baseline decreases, indicating the effectiveness of our approach in dealing with products with long context sequence. 

\subsection{Impact of Layer Mapping}
\label{sec:layer_mapping}

Layer mapping is employed to select the set of sparse layers in the heavy encoder to interact with, representing a key factor in sparse-layer interaction. To assess the impact of different layer mapping strategies, we conducted a series of experiments on both datasets using T5-large with 24 layers as the heavy encoder and T5-small with 8 layers as the light encoder. Various layer mapping schemes were explored on both datasets, and their effects were evaluated. These schemes include: 1) Even distribution layer, which maps every other 3 heavy encoder layers to a light encoder layer, starting with heavy encoder layers 0, 1, and 2. 2) Only using the last heavy encoder layer. 3) Using the last 8 heavy encoder layers. 4) Using the first 8 heavy encoder layers.
The values of $\alpha$ and $\beta$ were set to 0.7 and 1.0 on MAVE, and 0.05 and 0.0 on AE-110K.

Results are reported in Table \ref{tab:layer_mapping}. It can be observed that using heavy representations from the last layer achieves the best results on MAVE, while for AE-110K, the even distribution strategy generally performs better. The rationale behind this distinction is that, for MAVE, the heavy encoder is updated to provide improved task-specific representations in the last layer. In contrast, for AE-110K, the heavy encoder is frozen, and multiple layers contribute more diverse information.

\begin{table}[t]
\begin{adjustbox}{width=0.55\width,center}
\begin{tabular}{ccccccc}
\toprule
& \multicolumn{3}{c}{\textbf{MAVE}} & \multicolumn{3}{c}{\textbf{AE-110K}} \\
\midrule
\textbf{Layer mapping} & \textbf{$P$ (\%)} & \textbf{$R$ (\%)} & \textbf{$F_1$ (\%)}& \textbf{$P$ (\%)} & \textbf{$R$ (\%)} & \textbf{$F_1$ (\%)} \\
\midrule
{[}0, 3, 6, 9, 12, 15, 18, 21{]}           & 97.75           & 97.78           & 97.76       & \textbf{85.03}  & 84.21  & \textbf{84.62}      \\
{[}1, 4, 7, 10, 13, 16, 19, 22{]}          & 97.86           & 97.87           & 97.86       & 84.89  & 84.05  & 84.47     \\
{[}2, 5, 8, 11, 14, 17, 20, 23{]}          & 97.89           & 97.91           & 97.90       & 84.76  & 84.13  & 84.44     \\
{[}23, 23, 23, 23, 23, 23, 23, 23{]}       & \textbf{98.05}           & \textbf{98.09}           & \textbf{98.07}    & 83.92  & 84.29  & 84.10        \\
{[}16, 17, 18, 19, 20, 21, 22, 23{]}       & 98.05           & 97.98           & 98.02      & 84.36  & \textbf{84.36}  & 84.36      \\
{[}0, 1, 2, 3, 4, 5, 6, 7{]}               & 96.37           & 96.70           & 96.54   & 85.01  & 84.14  & 84.57    \\
\bottomrule
\end{tabular}
\end{adjustbox}
\vspace{-2mm}
\caption{Ablation study of layer mapping
on the MAVE and AE-110K datasets. Both are using EAVE L-S. For MAVE, we use $\alpha=0.7$ and $\beta=1.0$. For AE-110K, we use $\alpha=0.05$ and $\beta=0.0$.}
\label{tab:layer_mapping}
\vspace{-3mm}
\end{table}

\subsection{Representation Fusion Location}
\label{sec:fuse_location}
Within a Transformer encoder layer, the heavy and light representations can be extracted and fused at locations other than immediately after self-attention. Table~\ref{tab:fusion_location} presents a comparison of performance for six different fusion locations, including 1) Immediately before the self-attention layer and after its pre-normalization, 2) Immediately after the self-attention layer,
3) After the skip connection of the self-attention layer,
4) Immediately before the MLP layer after its pre-normalization,
5) Immediately after the MLP layer,
and 6) After the skip connection of the MLP layer.
As observed, for MAVE with $\beta=1.0$ (corresponding to fine-tuning the heavy encoder), the optimal fusion location is immediately before the self-attention layer and after its pre-normalization. For AE-110K with $\beta=0.0$ (corresponding to freezing the heavy encoder), the best fusion location is immediately before the MLP layer and after its pre-normalization.

\begin{table}[t]
\begin{adjustbox}{width=0.65\width,center}
\begin{tabular}{ccccccc}
\toprule
& \multicolumn{3}{c}{\textbf{MAVE}} & \multicolumn{3}{c}{\textbf{AE-110K}} \\
\midrule
\textbf{Fusion location}           & \multicolumn{1}{l}{$P$ (\%)} & \multicolumn{1}{l}{$R$ (\%)} & \multicolumn{1}{l}{$F_1$ (\%)} & \multicolumn{1}{l}{$P$ (\%)} & \multicolumn{1}{l}{$R$ (\%)} & \multicolumn{1}{l}{$F_1$ (\%)} \\
\midrule
before attn                    & \textbf{98.14}                      & \textbf{98.17}                      & \textbf{98.16}           & 84.11                      & 84.34                      & 84.23             \\
after attn                     & 97.95                               & 98.01                               & 97.98                      & 84.72                               & 84.08                               & 84.40           \\
after attn \& skip & 97.79                               & 98.05                               & 97.92                               & 85.04                               & 84.12                               & 84.58 \\
before mlp                          & 97.80                               & 97.91                               & 97.85                           & \textbf{85.08}                      & \textbf{84.39}                      & \textbf{84.74}        \\
after mlp                           & 97.68                               & 97.76                               & 97.72                           & 84.83                               & 84.12                               & 84.48     \\
after mlp \& skip       & 97.71                               & 97.94                               & 97.83                           & 84.44                               & 84.06                               & 84.25          \\
\bottomrule
\end{tabular}
\end{adjustbox}
\vspace{-2mm}
\caption{Ablation study on representation fusion location on both datasets. For MAVE, we use $\alpha=0.7$ and $\beta=1.0$. For AE-110K, we use $\alpha=0.05$ and $\beta=0.0$.}
\vspace{-3mm}
\label{tab:fusion_location}


\end{table}

\subsection{Heavy Encoder Size}
In this section, we ablate the heavy encoder size in our EAVE approach. The results on MAVE and AE-110K are shown in Table~\ref{tab:heavy_size}. As can be seen, for the MAVE dataset, as the heavy encoder size increases, the performance increases, while for AE-110K, as the heavy encoder size increases, the performance does not continuously increase. These results support our hypothesis that the context alone provides more useful information for the MAVE dataset comparing with the AE-110K dataset.

\begin{table}[h!]
\begin{adjustbox}{width=0.65\width,center}
\begin{tabular}{ccccccc}
\toprule
\textbf{}           & \multicolumn{3}{c}{\textbf{MAVE}}                               & \multicolumn{3}{c}{\textbf{AE-110K}}                   \\
\midrule
\textbf{Model size} & \textbf{$P$ (\%)} & \textbf{$R$ (\%)} & \textbf{$F_1$ (\%)} & \textbf{$P$ (\%)} & \textbf{$R$ (\%)} & \textbf{$F_1$ (\%)} \\
\midrule
B-S                 & 96.83           & 97.15           & 96.99              & 84.94           & 84.02           & 84.48              \\
L-S                 & 97.94           & 98.03           & 97.98              & 84.89           & 84.11           & 84.50              \\
XL-S                & 98.25  & 98.55  & 98.40     & 84.75           & 84.12           & 84.43      \\     
\bottomrule
\end{tabular}
\end{adjustbox}
\vspace{-2mm}
\caption{Ablation study on the heavy encoder size on both datasets. For MAVE, we use $\alpha=0.7$ and $\beta=1.0$. For AE-110K, we use $\alpha=0.05$ and $\beta=0.0$.}
\label{tab:heavy_size}
\vspace{-5mm}
\end{table}

\section{Conclusions}
Efficient product attribute value extraction is an important research problem in many real-world applications.
In this work, we proposed an efficient attribute value extraction method with lightweight sparse-layer interaction. Specifically, we decouple the computations of context-attribute non-interacting representations and interacting representations, using a heavy and light Transformer encoders respectively. Additionally, these representations are fused together through the sparse-layer interaction. We conducted benchmarks and systematic ablation studies on two open-sourced product attribute value extraction datasets. The results demonstrate that our method achieves much fast inference speed while maintaining the performance to the large encoder.

\section*{Limitations}
There are two limitations of our current EAVE approach. First, although the lightweight sparse-layer interaction method is tested in product attribute value extraction task, we believe this method can be generalized to any other tasks that are able to be converted to a context-query format, such as question answering, text pair classification, etc. As long as the context is long and number of queries per context is large, our method will offer dramatic efficiency gain. We plan to explore a generalization approach of our model. Second, we only investigated the lightweight sparse-layer interaction method under sequence labeling setup. However, recent works on Large language Models (LLM) have shown the success of decoder models. In the future, we plan to systematically investigate our method under Transformer decoder setup. 

\bibliography{anthology,custom}
\bibliographystyle{acl_natbib}

\clearpage
\newpage
\appendix

\section{Appendix}
\label{sec:appendix}

\subsection{Datasets}
\label{sec:datasets}
This section contains more details on the datasets. The train and eval splits on the selected attributes in Table~\ref{tab:mave_ae110_metrics} are shown in Table~\ref{tab:stats_selected_attributes_mave} for the MAVE dataset and Table~\ref{tab:stats_selected_attributes_ae110} for the AE-110K dataset. We select those attributes to ensure a broad spectrum of number of examples per attribute.
\begin{table}[h!]
\begin{adjustbox}{width=0.65\width,center}
\begin{tabular}{c|c|c|c|c|c}
\toprule
\multirow{2}{*}{Splits}          & \multicolumn{5}{c}{\textbf{MAVE}}                                        \\
\cmidrule{2-6}
 & \textbf{Season} & \textbf{Department} & \textbf{Resolution} & \textbf{Compatibility} & \textbf{All} \\
\midrule
Train              & 59199           & 20108               & 13792               & 6544                   & 4011570         \\
Eval               & 11431           & 3766                & 2492                & 1222                   & 755968      \\
\bottomrule
\end{tabular}
\end{adjustbox}
\caption{Statistics of the MAVE dataset.}
\label{tab:stats_selected_attributes_mave}

\quad

\begin{adjustbox}{width=0.7\width,center}
\begin{tabular}{c|c|c|c|c}
\toprule
\multirow{2}{*}{Splits}                                                                 & \multicolumn{4}{c}{\textbf{AE-110K}}                                           \\
\cmidrule{2-5}
 & \textbf{Brand Name} & \textbf{Material} & \textbf{Pattern Type} & \textbf{All} \\
\midrule
Train                    & 9098           & 3001                  & 1496                  & 88479        \\
Eval                     & 2316            & 810                  & 348                   & 21888       \\
\bottomrule
\end{tabular}
\end{adjustbox}
\caption{Statistics of the AE-110K dataset.}
\label{tab:stats_selected_attributes_ae110}
\end{table}

\subsubsection{MAVE}
MAVE is a dataset with long and structured context. After tokenization, the average context sequence length is 281, but there is a long tail distribution such that ~11.2\% examples are with sequence length larger than 512. We truncate or pad the context length to 512 and the attribute length to 32. Each product belongs to one category, which contains multiple possible attributes. During real-world application inference, for each product, we need to perform attribute value extraction for all attributes of its category. The statistics of several selected categories are shown in Table~\ref{tab:stats_selected_categories_mave}. As can be seen, some categories have O(10) number of attributes and a large amount of examples. Our EAVE method will achieve dramatic efficiency gain on such categories, which is presented in Section~\ref{sec:selected_categories}.

\begin{table}[h!]
\begin{adjustbox}{width=0.73\width,center}
\begin{tabular}{c|c|c|c|c}
\toprule
 Categories & \textbf{Shoes} & \textbf{Mobile Phones} & \textbf{Televisions} & \textbf{Dresses} \\
\midrule
Attributes     & 15             & 12                     & 11                                                  & 5                \\
\midrule
Train             & 599088         & 49543                  & 22041                                            & 99276            \\
Eval              & 114269         & 9671                   & 3958                                           & 18803      \\
\bottomrule     
\end{tabular}
\end{adjustbox}
\caption{Statistics of the MAVE dataset for the selected categories.}
\label{tab:stats_selected_categories_mave}
\end{table}

\subsubsection{AE-110K}
AE-110K is a dataset with context being only the title, so its context is short and simple. After tokenization, the average context sequence length is 32, and there are less than 0.1\% examples with sequence length larger than 64. We truncate or pad the context length to 64 and attribute length to 8. Different from the MAVE dataset, a product in AE-110K doesn't have a category label. But each product can contain multiple attributes. 

Following the work in \cite{aepub}, we manually fix several quality issues, including HTML entities, and extra white spaces in titles, attributes, and
values, and the same attributes sometimes have different letter cases. We thus decoded HTML entities,
converted trailing spaces into a single space, and removed white spaces at the beginning and ending.
We also remove duplicated tuples.

\subsection{Training Details}\label{training_details}
For all models in this paper, we use a learning rate of 1e-5, batch size 128, and Adam \cite{Kingma2014AdamAM} optimizer for training. We train up to 200k steps on 16 V5 Cloud TPUs using data parallelism. For both datasets, we use the same SentencePiece \cite{kudo-richardson-2018-sentencepiece} tokenizer as in the T5 paper \cite{2020t5}. We compute the GFlops numbers in Table~\ref{tab:mave_ae110_metrics} based on the sequence lengths mentioned in Section~\ref{sec:datasets}. More details on the hyper-parameters and configs are in Table~\ref{tab:hyperparams}.

Note that the learning rate, batch size and training steps are the same for all baselines, while the other hyperparameters are set to the optimal values in the original papers. W use the same learning rate, batch size and training steps for two main reasons. First, we want to ensure all models consume the same amount of training data (by using the same batch size and training steps) in order to achieve a fair comparison on the efficiency. Second, we set 200k training steps to ensure all models are sufficiently converged. In fact, we’ve observed that most methods converge within 50k steps. We also tried different hyperparameters and found the performances are quite stable.

\begin{table}[t]
\begin{adjustbox}{width=0.9\width,center}
\begin{tabular}{cc}
\toprule
Dropout Rate & 0.1 \\
Adam $\beta_1$ & 0.9 \\
Adam $\beta_2$ & 0.99 \\
Adam $\epsilon$ & 1e-8 \\
Activation Type & bfloat16 \\
\bottomrule
\end{tabular}
\end{adjustbox}
\caption{Hyper-parameters and configs.}
\label{tab:hyperparams}
\end{table}

\begin{table}[h!]
\begin{adjustbox}{width=0.65\width,center}
\centering
\begin{tabular}{ccccc}
\toprule
    & \multicolumn{2}{c}{\textbf{MAVE}} & \multicolumn{2}{c}{\textbf{AE-110K}} \\
\midrule
\textbf{Method}           & \textbf{$F_1$ (\%)} & \textbf{GFLOPS} & \textbf{$F_1$ (\%)} & \textbf{GFLOPS}  \\
\midrule
AVEQA (backbone) & 98.20                 & 402.47              & 85.54         & 16.11 \\
Pruning \cite{pruning} & 94.36                 & 246.62              & 80.17 & 8.85 \\
Distillation \cite{distillation} & 95.11                 & 270.49              & 81.55 & 9.23 \\
QAT \cite{QAT} & 93.57                 & 85.45              & 80.49 & 3.91 \\
EAVE (ours) & 97.98                 & 42.46              & 84.62 & 5.20 \\
\bottomrule
\end{tabular}
\end{adjustbox}
\caption{Comparison with traditional efficient methods, including Pruning \cite{pruning}, Distillation \cite{distillation} and Quantization \cite{QAT}, on both datasets.}
\label{tab:efficient_method}
\vspace{-5mm}
\end{table}
\subsection{Comparison with Traditional Efficient Techniques}
Traditional efficiency optimization techniques such as quantization, distillation and pruning are orthogonal to our method, as they are not targeting the efficiency improvement under the scenario of a single long context with multiple short queries. In this study, we conduct comprehensive experiments with these standard methods using AVEQA as the backbone. Specifically, for pruning \cite{pruning} technique, we prune 50\% of the network. For vanilla distillation \cite{distillation}, we distill to a 6-layer model with roughly 50\% parameters. For quantization, we use Quantization-Aware Training (QAT) \cite{QAT} with 4-bit from the original 32-bit. 

The results on both datasets are reported in Table \ref{tab:efficient_method}. It can be seen that while the GFLOPS of these methods decrease, there is significant performance drop compared with the original backbone. Moreover, when dealing with multiple attributes for a single product, they still need to encode the long product context multiple times. On the other hand, our approach is able to achieve good efficiency with on par performance, indicating the effectiveness of our modeling.

\begin{table}
\begin{adjustbox}{width=0.65\width,center}
\begin{tabular}{c|c|c|c|c}
\toprule
\multirow{2}{*}{Methods}                                                                 & \multicolumn{4}{c}{\textbf{AE-110K}}                                           \\
\cmidrule{2-5}
 & \textbf{Precision} & \textbf{Recall} & \textbf{F1} & \textbf{GFLOPS} \\
\midrule
SMARTAVE~\cite{smartave} & 85.12 &86.07 &85.49 &19.75\\
 LLaMa2 7B~\cite{llama2}  & 83.65 &84.77 &84.15 & 6578    \\
EAVE                     & 85.01  &84.24 &84.62 &5.20    \\
\bottomrule
\end{tabular}
\end{adjustbox}
\caption{Comparison results with LLaMa2 7B model on the AE-110K dataset.}
\label{tab:llm}
\end{table}
\begin{table}
\begin{adjustbox}{width=0.75\width,center}
\begin{tabular}{lccccc}
\toprule
\textbf{Category} & \textbf{Model} & \textbf{${\bf P}$ (\%)} & \textbf{${\bf R}$ (\%)} & \textbf{${\bf F_1}$ (\%)} & \textbf{GFLOPS} \\
\midrule
\multirow{4}{0.11\linewidth}{Shoes}   & T5-Small & 97.81 & 98.23 & 98.02 & 42.46 \\
&               T5-Base & 99.09 & 99.21 & 99.15 & 131.10 \\
&               T5-Large & 99.48 & 99.51 & 99.50 & 402.47 \\
&               EAVE L-S & 99.48 & 99.48 & 99.48 & 89.86 \\
\midrule
\multirow{4}{0.11\linewidth}{Mobile Phones}     & T5-Small & 81.02 & 89.81 & 85.19 & 42.46 \\
&               T5-Base & 90.86 & 95.57 & 93.16 & 131.10 \\
&               T5-Large & 95.04 & 97.57 & 96.29 & 402.47 \\
&               EAVE L-S & 96.08 & 96.95 & 96.52 & 96.22 \\
\midrule
\multirow{4}{0.11\linewidth}{Televisions}    & T5-Small & 88.32 & 92.57 & 90.39 & 42.46 \\
&               T5-Base & 95.81 & 97.06 & 96.43 & 131.10 \\
&               T5-Large & 98.37 & 98.61 & 98.49 & 402.47 \\
&               EAVE L-S & 98.69 & 98.53 & 98.61 & 99.11 \\
\midrule
\multirow{4}{0.11\linewidth}{Dresses}    & T5-Small & 94.53 & 95.5 & 95.01 & 42.46 \\
&               T5-Base & 97.75 & 97.98 & 97.87 & 131.10 \\
&               T5-Large & 98.78 & 98.87 & 98.83 & 402.47 \\
&               EAVE L-S & 98.85 & 98.79 & 98.82 & 140.72 \\
\bottomrule
\end{tabular}
\end{adjustbox}
\caption{Performance and cost comparison on the MAVE dataset sliced by categories with multiple attributes.}
\label{tab:mave_all_metrics}
\end{table}
\subsection{Comparison with LLM}\label{app:llm}
Large language models become the defacto in many NLP applications. Therefore, in this section, we conduct a comparison with LLaMa2 7B model~\cite{llama2}. For LLaMa2 7B, we set the batch size to 1, sequence length to 256, hidden dimension to 4096, number of layers to 32 and run it on 8 GPUs. For fair comparison between our method and LLM attribute value extraction, we extract different attributes separately by concatenating context and attribute as inputs to LLM. We also tried extracting attributes together in a single prompt, which leads to worse performance due to LLMs hallucination issue, and requires LLM with a larger finetuning set to achieve similar performance. The prompt we use is ``Please extract the attribute value of {attribute} from {context}''.

The evaluation results are reported in Table \ref{tab:llm}. As it can be seen in the table, LLaMa2 7B is able to achieve reasonable performance, while the inference cost is extremely expensive. Our hypothesis is that for the attribute value extraction task, in most cases, the value is from a text span in the product context and thus extractive models are more effective compared to the generative models. The observation is consistent with the findings in \cite{LLM1} that a smaller LLM like Beluga-7B utilizing in-context learning fails to outperform a fine-tuned BERT-based sequence tagging model (i.e., AVEQA) with a significantly smaller size, as indicated in their Tables 9 and 13.

\subsection{Results on Selected Categories}
\label{sec:selected_categories}
We present results on selected categories for the MAVE dataset in Table~\ref{tab:mave_all_metrics}. We show results of our EAVE model and three T5 baseline models of different sizes: Small, Base, and Large. GFLOPS per example is computed under a real world scenario: for a given product with $N$ attributes, we only need to compute context heavy representations once, so the GFLOPS per product is computed by $C^c_h / N + C^a_h + C_l + C_{sli}$. For simplicity, we don't consider the cost saving from caching heavy attribute representations. As can be seen from Table~\ref{tab:mave_all_metrics}, when $N$ increases we can observe the following: comparing with T5-Large, EAVE's performance is close, while its cost is dramatically reduced; comparing with T5-Small, EAVE's performance is significantly better, while its cost is only marginally heavier. The effectiveness of our approach on efficient is validated from those results.

\subsection{Practical Usage of Efficient Extraction}
In this section, we'd like to provide more insights into the importance and the practical usage of efficient attribute value extraction methods. 

First, it is a vast scale of real-world product system such as Amazon's or Google Shopping's online catalogs, numbering in the hundreds of millions to billions. It's important to note that each product is associated with multiple attributes, averaging more than 10 attributes per product. Consequently, any updates to the model or system necessitate a comprehensive re-extraction across all products, entailing billions of model inference calls (approximately ~100 million products multiplied by an average of 10 attributes per product). To illustrate, we conducted a test involving 2 billion extractions using a 3-layer distilled AVEQA model (BERT-base) across 20,000 CPU machines, which took more than two weeks to complete. 

Second, even if we were to limit the inference to only new and modified products, it would still require a significant time investment - approximately 7-10 hours per day. It's worth emphasizing that this scenario pertains to a 3-layer BERT-base model; the computational cost would escalate considerably for a full BERT model and LLMs. 

In addition, attribute value extraction serves as a crucial component for generating features in retrieval or ranking systems, such as web search and advertisements. For these applications, extraction must encompass all products available on the web across multiple platforms, potentially numbering in the tens of billions. Therefore, we believe that efficient product attribute value extraction represents a significant challenge.

\subsection{Extraction on Noisy Data}
In real-world product attribute value extraction, the products usually contain noise and even contradictory information within input texts. In fact, during our experimentation on product web pages extraction, we have indeed encountered failure cases due to noisy product data. For instance, a product title might mention `red’ shoes while the product description describes `blue’ pairs. One straightforward approach to address such discrepancies is to implement post-processing filtering similar to the method mentioned above. Specifically, when the model extracts multiple spans or values for an attribute (e.g., `red’ and `blue’) that do not align, we can simply return an UNKNOWN extraction.

\begin{table}[t]
\begin{adjustbox}{width=0.73\width,center}
\begin{tabular}{c|c|c|c|c}
\toprule
 Noise Level (during inference) & p = 0 & p = 0.1 & p = 0.2 & p = 0.4 \\
\midrule
trained on noisy data     & 96.96 & 97.10 & 96.99 & 97.25             \\
trained on clean data     & 97.27 & 95.36 & 93.57 & 90.39                \\
\bottomrule     
\end{tabular}
\end{adjustbox}
\caption{Model performance with noisy data.}
\label{noise}
\end{table}
To further understand the behavior of our model on noisy product data, we conduct an experiment by introducing noise to the context of the MAVE dataset. Specifically, with a probability $p$, for product description, we append up to 5 random selected attribute values from within the product category. With the same probability, we add another paragraph containing up to 5 random selected attribute values from within the product category. We then randomly split the noise augmented dataset into the train and eval set, and trained for 100k steps. The results are shown in Table \ref{noise}. It can be seen from the preliminary results that: 1) When training on noisy data, the performance of our model is relatively stable since the training also sees the noise while the label remains correct. 2) When training on clean data, the performance clearly drops, which is consistent with our expectation.

\end{document}